\documentclass[runningheads]{llncs}
\usepackage{graphicx}
\usepackage{subfigure}
\usepackage{amsmath}
\makeatletter
\newcommand{\printfnsymbol}[1]{%
  \textsuperscript{\@fnsymbol{#1}}%
}
\makeatother

\begin{document}
\title{A Combined Deep Learning-Gradient Boosting Machine Framework for Fluid Intelligence Prediction}
\titlerunning{A Combined DL-GBM Framework}
%
\author{Yeeleng S. Vang\inst{1}\thanks{equal contribution} \and
Yingxin Cao\inst{1}\printfnsymbol{1} \and
Xiaohui Xie\inst{1}}

%
\authorrunning{Y. Vang et al.}
%
\institute{University of California, Irvine CA 92697\\
\email{\{ysvang, yingxic4, xhx\}@uci.edu}}
\maketitle              
\begin{abstract}
The ABCD Neurocognitive Prediction Challenge is a community driven competition asking competitors to develop algorithms to predict fluid intelligence score from T1-w MRIs.  In this work, we propose a deep learning combined with gradient boosting machine framework to solve this task. We train a convolutional neural network to compress the high dimensional MRI data and learn meaningful image features by predicting the 123 continuous-valued derived data provided with each MRI.  These extracted features are then used to train a gradient boosting machine that predicts the residualized fluid intelligence score.  Our approach achieved mean square error (MSE) scores of 18.4374, 68.7868, and 96.1806 for the training, validation, and test set respectively.

\keywords{Fluid Intelligence  \and Deep Neural Network \and Machine Learning}
\end{abstract}
\section{Introduction}

The Adolescent Brain Cognitive Development (ABCD) study \cite{abcdstudy} is the largest long-term study of brain development and child health in the United States. Its stated goal is to determine how childhood experiences such as videogames, social media, sports, ect. along with the child's changing biology affects brain development. Understanding of brain development during the adolescent period is ``necessary to permit the distinction between premorbid vulnerabilities and consequences of behaviors such as substance use" \cite{luciana2018adolescent}. In this endeavor, leaders of the study organized the ABCD Neurocognitive Prediction Challenge (ABCD-NP-Challenge 2019) \cite{abcdchallenge} and invited teams to make predictions about fluid intelligence from T1-w magnetic resonance images (MRI). In psychology and neuroscience, the study of general intelligence often revolves around the concepts of fluid intelligence and crystallized intelligence.  Fluid intelligence is defined as the ability to reason and to solve new problems independent of previously acquired knowledge \cite{jaeggi2008improving} whereas crystallized intelligence is the ability to use skill and previous knowledge.  General intelligence is usually quantified by tests such as the Cattell Culture Fair Intelligence Test or Raven's Progressive Matrices.

This paper represents the authors' entry to the ABCD-NP-Challenge 2019 competition.  We propose a Convolutional Neural Networks (CNN) combined with gradient boosting machine (GBM) framework for the task of fluid intelligence prediction from 3D T1-w MRIs.  Our method combines the state-of-the-art approach of deep learning to find a good, non-linear compression of the high dimensional 3D MRI data and uses the superior performance of GBM to learn an ensemble regression model.

\section{Related Work}

The study of fluid intelligence has traditionally been more concern with trying to identify the underlying mechanism responsible for cognitive ability.  In \cite{Gray2003,cole2012global}, experiments were conducted to investigate factors and mechanisms that influence fluid intelligence with both suggesting that the lateral prefrontal cortex may play a critical role in controlling processes central to human intelligence. More recently, MRIs have been shown to contain useful structural information with strong correlation to fluid intelligence \cite{Colom2009}.  In the most related work to the subject of this paper, fluid intelligence was predicted directly from MRI data using support vector regressor obtaining an average corerleation coefficient of 0.684 and average root mean square error of 9.166 \cite{wang2015mri}, however the dataset consisted of only 164 subjects.  

In recent years, deep learning \cite{lecun2015deep} method has emerged as state-of-the-art solutions to many problems spanning various domain such as natural language processing, bioinformatics, and especially computer vision. Since winning the ImageNet competition in 2012 \cite{krizhevsky2012imagenet}, CNN, a type of deep learning model, have been the defacto tool for analyzing image data. Its impact in the biomedical image domain has been nothing short of extraordinary \cite{shen2017deep}.  Many disease, such as cancer, that are detectable and segmentable by radiologists studying brain MRI can now be automatically performed by deep learning algorithms  \cite{akkus2017deep,havaei2017brain} with performances now comparable to many experts \cite{tang2018automated,zhu2018anatomynet}.

\section{Model}

Our model pipeline consists of two parts.  As these 3D MRIs are of very high dimensions, we first train a CNN model to learn a data compression scheme that best preserves meaningful features of the original image data.  Using this CNN, we feature extract a compressed version of the original input image to better utilize limited computer memory. These compressed images are used as surrogates for the original images and is used to train the GBM to learn a prediction model. These two steps are explain in further details in the following sections.

\subsection{Feature Extractor}

\begin{figure}[!h]
\includegraphics[width=\textwidth]{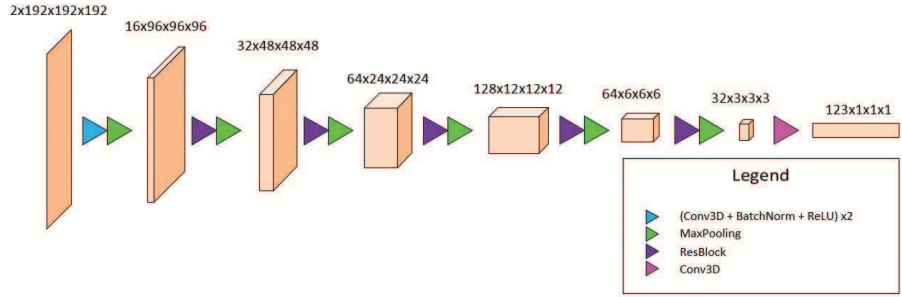}
\caption{Illustration of feature extractor model.  This is a fully convolutional network designed in an encoder architecture for easy access to data compression at different image resolutions.} \label{fig1}
\end{figure}

The original T1-w MRIs are much too high dimensional to be used with many modern machine learning algorithms with more than a few samples at a time due to computing memory limitations. This necessitates first compressing these original MRIs to a more compact representation while preserving meaningful features before learning a regression model.  Another motivation for using a compact representation is to reduce potential overfitting.  In our proposed framework, we use a fully convolutional network (FCN) to encode the raw MRIs to a compressed form.  It is fully convolutional because we only use the convolution and pooling operations. The structure of the CNN is shown in Fig. \ref{fig1}. 

The dual-channel input to our FCN consists of the original T1-w MRI and segmented brain MRI.  These two images are cropped into $192^3$-voxel cubes and passed through our FCN resulting in a real-valued vector output of length 123. The length of this one-dimensional vector is design to match the number of derived, continuous covariates for each MRI \cite{pfefferbaum2017altered}. These derived data are discussed in more detail in the data section. We formulate the training of the feature extractor as a regression problem calculating the mean square error (MSE) between predicted output and the continuous covariates as the loss function. The motivation to use this approach is due to the fact that deep learning models have trouble predicting a single continuous value when the input is of such high dimensions. As the 123 continuous covariates are unqiue for each subjects, the thought is that our model will learn better features for downstream task.  We also design the FCN in an encoder fashion to permit easy extracting of features at different image scales. 

The model is trained for 15 epochs with a learning rate of 0.01, momentum of 0.9, and batch size of 4. The stochastic gradient descent with momentum optimizer is used.  The model is trained on both the training and test datasets and validated on the validation set.  The epoch with the lowest validation set MSE score was used to extract features.  We extracted features at both the $6\times6\times6$ and $3\times3\times3$ scales.  We found empirically that the $6\times6\times6$ scale produced better results in the latter fluid intelligence prediction part.

\subsection{Gradient Boosting Machine}
Following feature extraction, the extracted images are regressed upon using gradient boosting machine (GBM) \cite{friedman2001greedy}.  GBM is a boosting method that obtains a strong predictor by ensembling many weak predictors. It iteratively improves its current model by adding a new estimator fitted to the residual of the current model and true labels. Specifically, GBM learns a functional mapping $\hat{y}=F(x;\beta)$ from data $\{x_i,y_i\}_{i=1}^N$ that minimizes some loss function $L(y,F(x;\beta))$.  More concretely, $F(x)$ takes the form of $F(x)= \sum_{m=0}^M \rho_m f(x;\tau_m)$ where $\rho$ is the weight, $f$ is the weak learner, and $\tau$ is the parameter set. Then $\beta$ consists of the parameter sets $\{\rho_m, \tau_m\}$.  These parameter sets are learned in the following stage-wise greedy process:

\{1\} set an initial estimator $f_0 (x)$.

\{2\} for $m \in 1,2,\dots, M$
\begin{equation}
(\rho_m, \tau_m) = \operatorname*{arg\,min}_{\rho,\tau} \sum_{i=1}^n L(y_i,F_{m-1}(x_i) + \rho f(x_i;\tau))
\end{equation}

Step \{2\} is approximated by GBM in the following two steps:

First, learn $\tau$ by:
\begin{equation}
\tau_m = \operatorname*{arg\,min} \sum_{i=1}^n (g_{im} - f(x_i;\tau))^2
\end{equation}

where $g_{im}=-\left[\frac{\partial L(y_i, F(x_i))}{\partial F(x_i)} \right]_{F(x)=F_{m-1}(x)}$

Second, learn $\rho$ by: 
\begin{align*}
\rho_m = \operatorname*{arg\,min}_{\rho} \sum_{i=1}^n L(y_i,(F_{m-1}(x_i) + \rho f(x_i;\tau_m))
\end{align*}

Finally it updates $F_m(x) = F_{m-1}(x) + \rho f(x;\tau)$.  To control overfitting, shrinking is introduced to give the following form of the update equation: $F_m(x) = F_{m-1}(x) + \gamma \rho f(x;\tau)$, where $0 \leq \gamma \leq 1$. The performance of GBM is improved by random subsampling the training data to fit each new estimator \cite{friedman2002stochastic}. 

In this project, we use the XGBoost \cite{chen2016xgboost} implementation of GBM for fast GPU support using trees. The GBM model requires a number of hyperparameters to be set.  We perform a two stage grid search to find the best combination of hyperparameters.  The first stage involves searching over a coarse grid to obtain the general vicinity of the optimal hyperparameters.  The second stage involves searching over a fine grid around the best hyperparameter from the first stage.  The following hyperparameters are found to give the best validation set mean square error score:  Learning rate = 0.006, number of trees = 1000, depth of tree = 7.

To minimize overfitting, we use elastic net regularization \cite{zou2005regularization} which adds the LASSO and RIDGE regularization to Eqn (1) and Eqn (2). Specifically, $\frac{1}{2} \lambda \sum_{m=1}^{M} \rho^2 + \alpha \sum_{m=1}^{M} |\rho|$ is added to Eqn (1) and $\frac{1}{2} \lambda \sum_{m=1}^{M} \tau^2 + \alpha \sum_{m=1}^{M} |\tau|$ is added to Eqn(2). The elastic net regularization parameters are tuned against the validation dataset and set as $\lambda=1.05$ and $\alpha=0.1$ to achieve the results of Table \ref{tab1}.

\section{Experiment}
\subsection{Data}

\begin{figure}%
\centering
\subfigure[]{%
\label{fig2a}%
\includegraphics[height=2.2in]{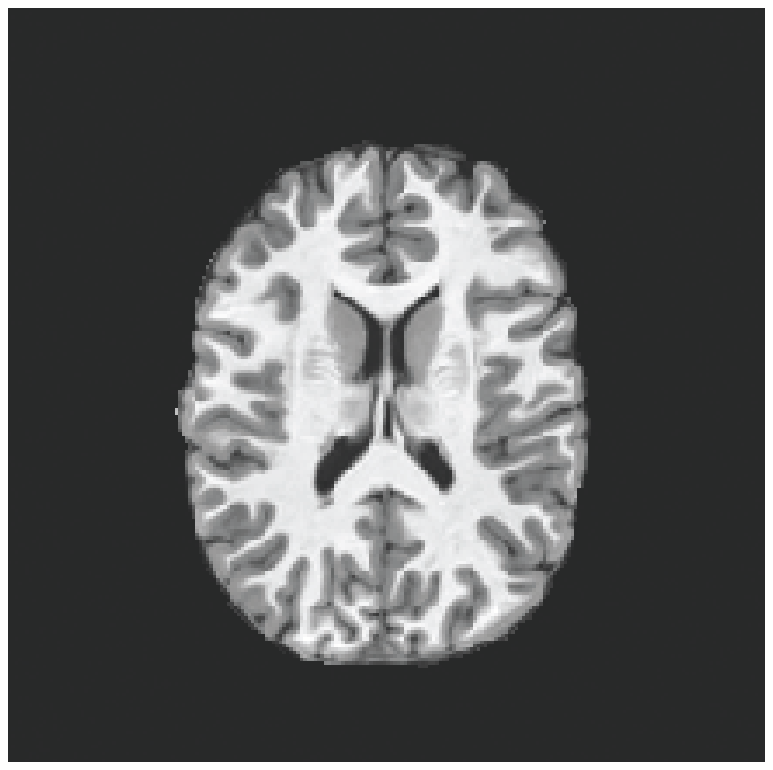}}%
\qquad
\subfigure[]{%
\label{fig2b}%
\includegraphics[height=2.2in]{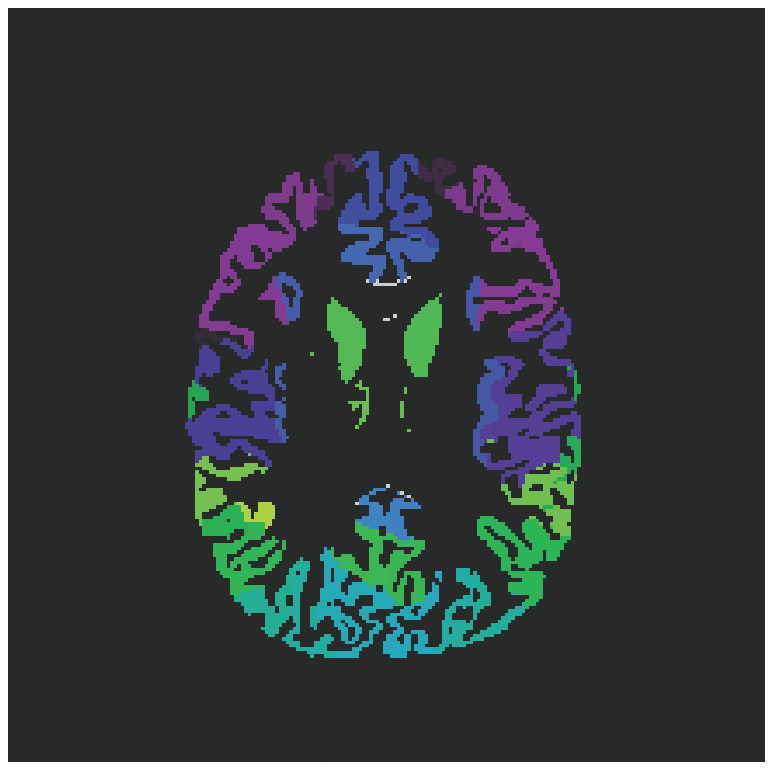}}%
\caption{(a) Example of pre-processed T1-w brain MRI. (b) Example of the white matter, gray matter, and CSF segmented brain MRI of the same T1-w image. }\label{fig2}
\end{figure}

3D T1-w MRIs were pre-processed and provided by the challenge organizer.  The pre-processing steps involved first creating brain masks from a series of standard brain extraction softwares including FSL BET, AFNI 3dSkullStrip, FreeSurer mri\_gcut, and Robust Brain Extraction (ROBEX).  The final brain mask was obtained by taking majority voting across these resulting masks.  This final brain mask was used to perform bias correction and the extracted brain was segmented into brain tissue (gray matter, white matter, and cerebrospional fluid (CSF)) using Atropos.  Finally the skull-stripped brain and segmented brain images were registered affinely to the SRI24 atlas \cite{pfefferbaum2017altered}.  Fig. \ref{fig2} shows typical example of the pre-processed, skullstripped T1-w MRI and segmented brain MRI provided by the organizers.

The fluid intelligence scores were pre-residualized by the challenge organizer.  It involved fitting a linear regression model to fluid intelligence as a function of the following covariates: brain volume, data collection site, age at baseline, sex at birth, race/ethnicity, highest parental education, parental income, and parental marital status.  Subjects in the ABCD NDA Release 1.1 dataset missing any value was excluded from the training and validation set. Once the linear regression model was fitted, fluid intelligence residual for each patients were calculated and constitutes the value to predict for this competition.

In addition to the provided 3D T1-w MRI, segmented brain MRI, and the fluid intelligence residual scores, derived data \cite{pfefferbaum2017altered} was provided for each patient.  These derived data consisted of 123 continuous-valued, volumetric data for different gray and white matter tissues such as ``right precentral gyrus gray matter volume" and ``corpus callosum white matter volume".  In all, the training set consists of 3739 patients, validation set consists of 415 patient, and test set consists of 4402 patients.

\subsection{Evaluation Criteria}

The competition evaluation criteria is based on the mean square error (MSE) score between predicted and true residualized fluid intelligence score.  This is calculated as the following equation:

\begin{equation*}
MSE = \frac{1}{N} \sum_{i=1}^N (y - \hat{y})^2
\end{equation*}
where $N$ is the total number of subjects in each data fold, $y$ is the true residualized fluid intelligence score, $\hat{y}$ is the predicted score from our model.
The organizers provided an evaluation script written in the R language to perform this evaluation for the training and validation set for consistency between competitors.  Ranking is provided separately for both validation set performance and test set performance.  

\begin{table}[]
\centering
\caption{Proposed Network MSE performance}\label{tab1}
            
\begin{tabular}{cccl}
                     & Train Set & Validation Set & Test Set \\ \hline
BrainHackWAW         & ---       & 67.3891        & 92.9277  \\ \hline
MLPsych              & ---       & 68.6100        & 95.6304  \\ \hline
CNN+GBM (Our method) & 18.4374   & 68.7868        & 96.1806  \\ \hline
BIGS2                & ---       & 69.3861        & 93.1559  \\ \hline
UCL CMIC             & ---       & 69.7204        & 92.1298  \\ \hline
\end{tabular}
\end{table}

\subsection{Results}

The result of our algorithm is reported in Table \ref{tab1} along with the other top four performing teams based on MSE ranking of the validation set performance at the end of the competition.  Our approach achieved a third place finish on the validation set.  The best performing team based on the test set was 4.3\% better than our results.

Table \ref{tab2} shows the ablation study proving the value of incorporating MRI date for fluid intelligence prediction.  The baseline ``derived data+GBM" learns a GBM regression model on the derived data directly.  Using our proposed method incorporating T1-w MRI data reduces training set MSE by 25.5\% and validation set MSE by 5.4\%.

\begin{table}[]
\centering
\caption{Ablation study}\label{tab2}
\begin{tabular}{lll}
                    & \multicolumn{1}{c}{Training Set MSE} & \multicolumn{1}{c}{Validation Set MSE} \\ \hline
Derived Data+GBM & \multicolumn{1}{c}{24.7683}          & \multicolumn{1}{c}{72.7308}            \\ \hline             
CNN+GBM & \multicolumn{1}{c}{18.4374}          & \multicolumn{1}{c}{68.7868}            \\ \hline
                    &                                      &                                        \\
            
\end{tabular}
\end{table}

\section{Conclusion}

In this paper we propose a combination deep learning plus gradient boosting machine framework for the task of fluid intelligence prediction based on T1-w MRI data.  Specifically we design a fully convolutional network to perform data compression/feature extraction.  Using these extracted features, we employ gradient boosting machine to learn an ensemble model for regression.  Our model achieved a third place finish based on the validation set ranking.

For future directions, we would like to investigate different encoding scheme such as autoencoders to perform the feature extraction step. In theory, autoencoders are able to find the optimal compression scheme which may improve the downstream regression task. About the regression model, instead of using elastic net as the regularization technique, we can consider replacing them with the more recently developed dropout \cite{rashmi2015dart} to minimize the gap in training set and validation set MSE performance for better generalization.

%
%
%
\bibliographystyle{splncs04}
\bibliography{reference}

\end{document}